\documentclass[letterpaper, 10 pt, conference]{ieeeconf}  

\IEEEoverridecommandlockouts                              

\usepackage{graphics} 
\usepackage{epsfig} 
\usepackage{times} 
\usepackage{amsmath} 
\usepackage{amssymb}  
\usepackage{graphicx}
\usepackage{array}
\usepackage{hyperref}
\usepackage[algoruled, resetcount, linesnumbered ]{algorithm2e}
\usepackage{multirow}
\usepackage[dvipsnames]{xcolor}
\usepackage{pifont}
\usepackage{bm}
\usepackage{mathtools}
\usepackage{lipsum}
\usepackage[english]{babel}
\usepackage{blindtext}
\usepackage[capitalize]{cleveref}
\usepackage{siunitx}
\usepackage{MnSymbol}  
\usepackage{color,soul}

\graphicspath{{./figs/}}
\newcolumntype{C}[1]{>{\centering\arraybackslash}p{#1}}

\newcommand{\cmark}{\ding{51}}%

\let\oldnl\nl
\newcommand{\nonl}{\renewcommand{\nl}{\let\nl\oldnl}}
\makeatletter 
\g@addto@macro{\@algocf@init}{\SetKwInOut{Parameter}{Parameters}} 
\makeatother

\newcommand{\doctitle}{
Under the Radar: Learning to Predict Robust Keypoints for \\Odometry Estimation and Metric Localisation in Radar
}
\newcommand{\docsubtitle}{}
\title{\LARGE \bf\doctitle\docsubtitle} 
\author{Dan Barnes and Ingmar Posner
\thanks{Authors are from the Applied AI Lab, University of Oxford, UK.
{\tt\small \{dbarnes,ingmar\}@robots.ox.ac.uk}}
}

\begin{document}

\maketitle
    \thispagestyle{empty}
\pagestyle{empty}

\begin{abstract}

This paper presents a self-supervised framework for learning to detect robust keypoints for odometry estimation and metric localisation in radar.
By embedding a differentiable point-based motion estimator inside our architecture, we learn keypoint locations, scores and descriptors from localisation error alone.
This approach avoids imposing any assumption on what makes a robust keypoint and crucially allows them to be optimised for our application. 
Furthermore the architecture is sensor agnostic and can be applied to most modalities.
We run experiments on 280km of real world driving from the Oxford Radar RobotCar Dataset and improve on the state-of-the-art in point-based radar odometry, reducing errors by up to 45\% whilst running an order of magnitude faster, simultaneously solving metric loop closures. Combining these outputs, we provide a framework capable of full mapping and localisation with radar in urban environments. 

\end{abstract}


\section{Introduction}

Robust egomotion estimation and localisation are critical components for autonomous vehicles to operate safely in urban environments. 
Keypoints are routinely used in these applications but are typically manually designed or not optimised for the task at hand. 
Keypoints represent repeatable locations under different viewpoints; hence conditions such as lighting (lens-flare), weather (rain) or time of day (night) can have drastic effects on their quality.
The key to improving keypoint quality and robustness is to learn keypoints \emph{specifically tailored} for these tasks and sensor modality.

There is increasing research into radar for urban robotics and because of it's wavelength and range holds the promise of directly addressing many of aforementioned challenges. 
However, it is also a notoriously challenging sensing modality: typically covered with noise artefacts such as ghost objects, phase noise, speckle and saturation.
Hence using off-the-shelf keypoint detectors designed for other modalities is ill-advised, but makes radar an ideal, if challenging, candidate for learning a more optimal keypoint detector. 

In this paper we present a self-supervised approach for learning to predict keypoint locations, scores and descriptors in radar data for odometry estimation and localisation. 
We achieve this by embedding a differentiable point-based motion estimator inside our architecture and supervise only with automatically generated ground truth pose information.
This approach avoids imposing any assumption on what makes a robust keypoint; crucially allowing them to be optimised for our application rather than on some proxy task. 
Furthermore, the architecture itself is sensor agnostic as long as real world keypoint locations can be inferred. 

Our approach leads to a state-of-the-art in point-based radar odometry when evaluated on the Oxford Radar RobotCar Dataset \cite{RadarRobotCarDatasetArXiv} driving in complex urban environments. 
In addition the formulation detects metric loop closures, leading to a full mapping and localisation system in radar data.

\begin{figure}
    \centering
    \includegraphics[width=\linewidth]{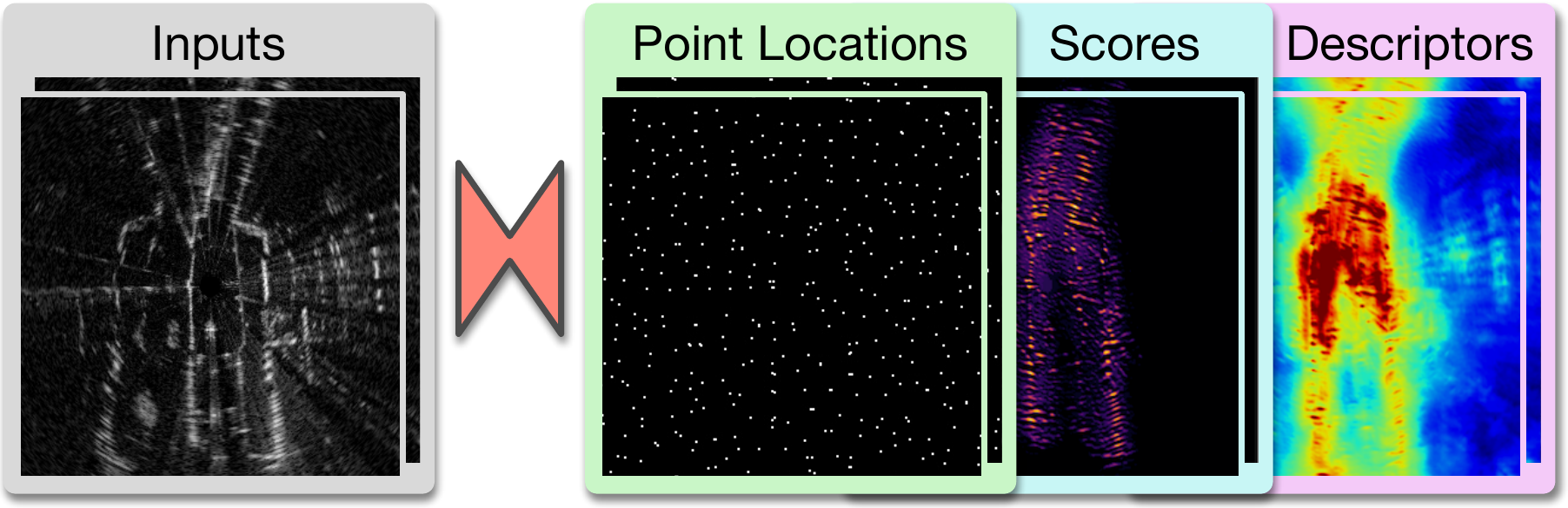}\\[0.4em]
    \includegraphics[width=\linewidth,clip,trim={ 0 1.6cm 0cm .8cm }]{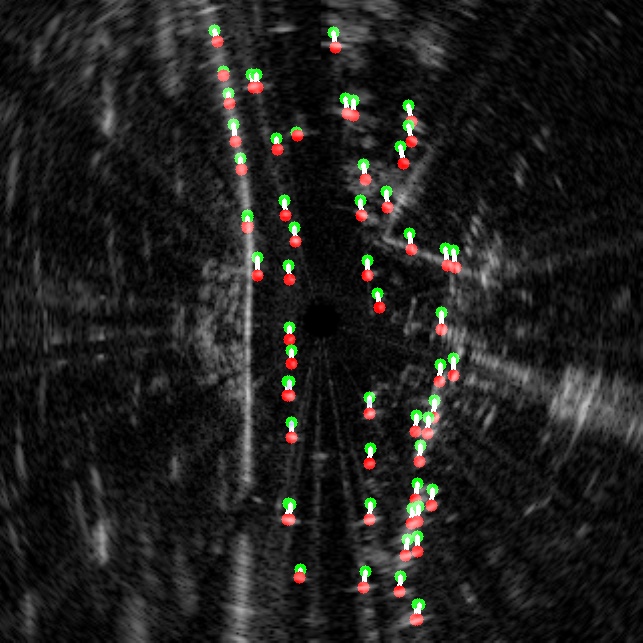}
    \caption{
    Learned keypoints for localisation in radar. 
    Given pair of input radar scans (top left) a trained CNN predicts keypoint locations, scores and descriptors (top right).
    We calculate point matches using descriptor cosine similarity; with final matches also weighted by keypoint scores allowing us to ignore points belonging to noise and unobservable regions (centre with only highest scoring points shown).
    Finally a pose estimator calculates the optimal transform from the point matches.
    Crucially the formulation is fully differentiable and can be supervised on odometry error alone, thereby learning keypoint locations, scores and descriptors that are optimal for localisation.
    }
    \vspace{-.375cm}
    \label{fig:intro_fig}
\end{figure}


\begin{figure*}
    \centering
    \includegraphics[width=\linewidth]{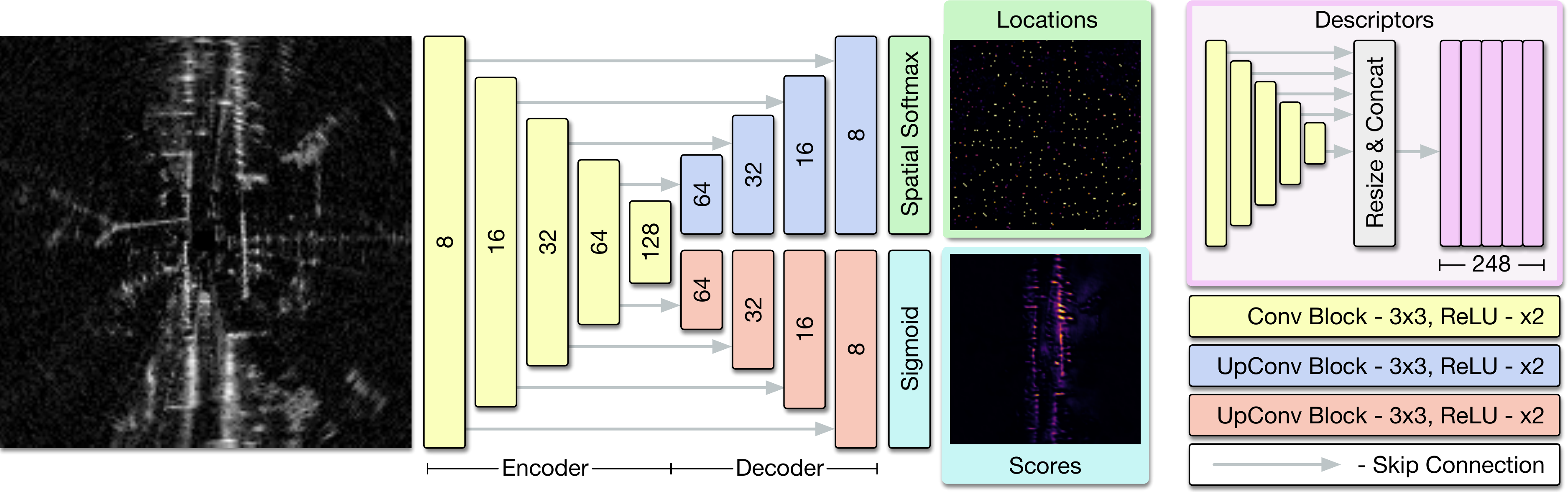}
    \caption{
    Network architecture for predicting keypoint locations, scores and descriptors. 
    The height of each block indicates the spatial dimensions of the feature map, which vary by a factor of 2 between blocks through max-pooling or bilinear-interpolation. 
    A dense pixel-wise descriptor map (top right) is created by resizing the output of each encoder block to the size of the input before concatenation into a single feature map. 
    For keypoint locations, spatial softmax is performed on a per cell basis with cell size chosen such that 400 keypoints are predicted.
    A pointwise convolution with no activation and single channel output precedes the sigmoid and spatial softmax operations. The number of output channels are detailed for each block.
    }
    \vspace{-.35cm}
    \label{fig:network_architecture}
\end{figure*}

\section{Related Work}

Extracting keypoints, such as SIFT \cite{Lowe:1999:ORL:850924.851523}, SURF \cite{bay2006surf} and ORB \cite{rublee2011orb},  from sensor data has historically been an initial step for egomotion estimation, place recognition, and simultaneous localisation and mapping (SLAM).
Recently CNN based keypoint detectors have emerged predicting locations \cite{detone2017toward,laguna2019key} and also descriptors \cite{detone2018superpoint,yi2016lift}. 
However ground-truth supervision is challenging as any reliably detected location is a candidate keypoint. 
Typically, these keypoint detectors are trained with homography related losses to promote keypoint repeatability or use labels from other detectors. 
However, both these solutions are suboptimal given the alternative of learning keypoints tailored for a downstream task.

By embedding a differentiable point-based pose estimator \cite{suwajanakorn2018discovery} learns to predict keypoint locations for the task of rotation prediction; however the formulation predicts only object category specific keypoints which cannot generalise to new scenes.
Conversely \cite{wang2019deep} registers two point clouds by predicting point-wise descriptors for matching, followed by the same pose estimation formulation.

Our target domain of radar is becoming an increasingly researched modality for mobile robotics and with the recently released Oxford Radar RobotCar Dataset \cite{RadarRobotCarDatasetArXiv}, a radar extension to the Oxford RobotCar Dataset \cite{RobotCarDatasetIJRR}, we expect interest to continue to grow.
The seminal work on egomotion estimation in this modality \cite{cen2018precise} extracts point features before predicting pose; 
however the point extraction is hand-crafted and may not be optimal for the task.
The current state-of-the-art in egomotion estimation in radar \cite{MaskingByMovingArXiv} employs a correlation-based approach, 
with learned masking to ignore moving objects and noise artefacts, but is limited to a maximum rotation, making it unsuitable for metric localisation.

Inspired by \cite{suwajanakorn2018discovery, wang2019deep} we learn to predict keypoints specialised for localisation by embedding a pose estimator in our architecture and use only pose information as supervision. 
This avoids imposing any assumptions on what makes suitable keypoints and enables pose prediction at any angle, a limitation of the current state-of-the-art \cite{MaskingByMovingArXiv}.
Furthermore training over a large dataset, we produce descriptors ideally suited for place recognition without tailored architectures \cite{arandjelovic2016netvlad} or training regimes designed for that task.

\section{Learning Point-Based Localisation}\label{sec:learning}
In the following section we outline our approach for learning roboust keypoints from ground truth pose information. 
No part of the approach or model design have been tailored for radar data and can be applied to other modalities such as vision or LIDAR.
Our method takes the following steps:

\emph{1) Keypoint Prediction:} From a raw radar scan we predict keypoint locations, scores and descriptors. 

\emph{2) Pose Estimation:} Given keypoints from two proximal scans we estimate the optimal transform between them and use the errors to train keypoint prediction.

\emph{3) Metric Localisation:} Using the same keypoint descriptors as a summary of the local scene, we detect and solve metric loop closures.

\subsection{Keypoint Prediction}\label{sec:learning:keypoint_prediction}
\vspace{-0.075em}
We adopt a U-Net \cite{ronneberger2015u} style convolutional encoder-multi-decoder network architecture (with concatenation skip connections) as shown in \cref{fig:network_architecture} to predict full resolution point locations, scores and descriptors. 

The Locations head predicts the sub-pixel locations of each keypoint. 
To achieve this, we divide the full resolution Locations output into equally sized square cells, with each producing a single candidate keypoint. 
We apply a spatial softmax on each cell followed by a weighted sum of pixel coordinates to return the sub-pixel keypoint locations. 

The Scores head predicts how useful a keypoint is for estimating motion and is mapped to $[0, 1]$ by passing the full resolution logits through a sigmoid function. A perfect scores output would give all static structure in the scene, such as walls and buildings, a score of $1$ and all noise, moving objects and empty regions a score of $0$.

The Descriptors aim to uniquely identify real-world locations under keypoints so that we can relate points by comparing descriptor similarity.
Dense descriptors are created by resizing the output of each encoder block (shown in yellow) to the input resolution before concatenation into a single 248 channel feature map. 

\subsection{Pose Estimation}\label{sec:learning:pose_estimation}
\vspace{-0.075em}
Given a set of keypoint locations we can extract keypoint descriptors and scores using $@$, where $@$ is a sampling function so that $X \; @ \; y$ takes a bilinear interpolation of dense feature map $X$ at coordinates $y$. 
The keypoint descriptors are then $\ell_2$ normalised so that cosine similarities between any pair is in the range $[-1, 1]$ using: $\tilde{d} = \ell_2(d) = d \; / \; ||d||_2$.

Given two proximal radar scans and their predicted keypoint locations, scores and descriptors we can match keypoints using the differentiable formulation in \cref{alg:point_matching} producing keypoint matches ($\bm{P}_s, \bm{P}_d$) and weights ($\bm{w}$) in the range $[0, 1]$. 
The weights are a combination of the keypoint scores and descriptor cosine similarity; hence matches are only kept if part of the static scene, as predicted by keypoints scores, and identified the same real world location by comparing keypoint descriptors. 

The matching is implemented as a dense search for optimum keypoint locations in the destination radar scan given keypoints in the source radar scan. 
Although matching keypoints directly would be computationally preferable, this formulation produces improved results while still running at well over real-time speeds.

Similar to \cite{suwajanakorn2018discovery, wang2019deep} given matched keypoints and weights we calculate the transform between them using singular value decomposition (SVD) as laid out in \cref{alg:pose_estimation} (detailed further in \cite{svd_pose_estimation_notes}).
Crucially this pose estimation is differentiable, allowing us to backpropagate from transform error right through to the keypoint prediction network. 

Given a ground-truth transform between two radar scans, we train with a loss penalising the error in translation and rotation, with weight $\alpha = 10$, learning keypoints optimal for motion estimation.
\begin{equation}
\mathcal{L} = ||\hat{t} - t||_2 + \alpha\;||\hat{R}R^{\mathsf{T}} - \mathbf{I}||_2
\end{equation}

\begin{algorithm}
\caption{Differentiable Point Matching}\label{alg:point_matching}
\KwIn{}
\nonl \hspace{.25em} $\bm{P}_s$ \hspace{1.875em} // source point pixel locations \\
\nonl \hspace{.25em} $\bm{D}_s$, $\bm{D}_d$ // source and destination descriptor maps \\
\nonl \hspace{.25em} $\bm{S}_s$, $\bm{S}_d$ \hspace{0.125em} // source and destination score maps \\
\Parameter{}
\nonl \hspace{.25em} $T$ \hspace{.0002em} // descriptor cosine distance softmax temperature \\
\nonl \hspace{.25em} $\bm{X}$ // pixel locations map\\
\KwOut{}
\nonl \hspace{.25em} $\bm{P}_d$ // destination point locations \\
\nonl \hspace{.25em} $\bm{w}$ \hspace{.0002em} // point match weights \\
\SetAlgoLined
\BlankLine

\For{$i\gets1$ \KwTo $n$}{

    \nonl \vspace{-1.05em}\hspace{7em} // For each source point \vspace{0.05em} \\

    \nonl // Extract and normalise source point descriptor \\
    $\bm{d}_{si} \leftarrow \ell_2\;(\; \bm{D}_s @ \; \bm{p}_{si} \; )$ \\

    \nonl // Pixelwise cosine distance to dest. descriptor map \\
    $\bm{C}_i \leftarrow \bm{d}_{si} \odot \bm{D}_d$ \\
    
    \nonl // Apply temperature weighted softmax \\
    $\bm{S} \leftarrow \sigma ( T \bm{C}_i)$ \\
    
    \nonl // Extract destination point pixel coordinates \\ 
    $\bm{p}_{di} \leftarrow \bm{S} \odot \bm{X}$ \\
    
    \nonl // Extract and normalise destination point descriptor \\
    $\bm{d}_{di} \leftarrow \ell_2\;(\; \bm{D}_d @ \; \bm{p}_{di} \; ) $ \\
    
    \nonl // Extract source and destination point scores \\
    $s_{si} \leftarrow \bm{S}_s @ \; \bm{p}_{si} \, , \; s_{di} \leftarrow \bm{S}_d @ \; \bm{p}_{di} $ \\
    
    \nonl // Compute weight for point match \\
    $w_i \leftarrow \frac{1}{2}(\bm{d}_{si} \odot \bm{d}_{di} + 1) ~ s_{si} ~ s_{di} $ \\
}
\end{algorithm}

\begin{algorithm}

\caption{Differentiable Pose Estimation}\label{alg:pose_estimation}
\KwIn{}
\nonl \hspace{.25em} $\bm{P}_s$, $\bm{P}_d$ // source and destination point pixel locations \\
\nonl \hspace{.25em} $\bm{w}$ \hspace{1.7202em} // point match weights \\
\KwOut{}
\nonl \hspace{.25em} $\bm{t}$, $R$ // optimal translation and rotation that minimise: \\
{\centering
\nonl \hfill $\sum_{i=1}^n w_i \| (R \bm{q}_{si} + \bm{t}) - \bm{q}_{di} \|^2 $\hfill
}
\BlankLine
\nonl // Convert pixel locations to world locations \\
$\bm{Q}_s \leftarrow \text{pix2world}(\bm{P}_s) \, ,  \; \bm{Q}_d \leftarrow \text{pix2world}(\bm{P}_d)$ \\
\BlankLine
\nonl // Compute the weighted centroids of both point sets \\
\vspace{0.25em}
$\bar{\bm{Q}_s} \leftarrow \sum_{i=1}^{n} w_i \; \bm{q}_{si}  \; /  \; \sum_{i=1}^{n} w_i$ \\
\vspace{0.25em}
$\bar{\bm{Q}_d} \leftarrow \sum_{i=1}^{n} w_i \; \bm{q}_{di}  \; /  \; \sum_{i=1}^{n} w_i$ \\
\BlankLine
\nonl // Compute the centred vectors \\
$\bm{x}_i \leftarrow \bm{q}_{si} - \bar{\bm{Q}_s} \, ,  \; \bm{y}_i \leftarrow \bm{q}_{di} - \bar{\bm{Q}_d} \, ,  ~~~~ i = 1,2,...,n.$
\BlankLine
\nonl // Compute the d x d covariance matrix \\
$S \leftarrow XWY^T$ \\
\BlankLine
\nonl // Compute the singular value decomposition \\
$U, {\mathrel{\raisebox{1.6pt}{$\scriptstyle\sum$}}}, V \leftarrow SVD(S)$ \\
\BlankLine
\nonl // Compute optimal rotation \\
$R \leftarrow V \begin{psmallmatrix}
    1 & & & \\
    & 1 & & \\
    & & \ddots & \\
    & & & det(VU^T) \\
  \end{psmallmatrix} U^T$  \\
\BlankLine
\nonl // Compute optimal translation \\
$\bm{t} \leftarrow \bar{\bm{Q}_d} - R \bar{\bm{Q}_s}$ \\
\end{algorithm}

\subsection{Metric Localisation}\label{sec:learning:localisation}

For each radar scan we assemble a dense descriptor map which enables the keypoint matching previously discussed. 
Although trained for the task of pose estimation, we reuse the descriptors to produce a location specific embedding $\bm{G}$ by max-pooling the dense descriptors $\bm{D}$ across all spatial dimensions, resulting in a single 248-D embedding.
This process adds practically no overhead to the inference speed of the network. 
At run-time, we compare the cosine similarity of the current embedding to previously collected embeddings; when the similarity crosses a threshold, the pair is deemed to be a topological loop closure (at the same physical location).
When a topological loop closure is detected, the respective keypoints are solved for a full metric loop closure using the same pose estimation formulation detailed in \cref{sec:learning:pose_estimation}.

\section{Experimental Setup}

We aim to evaluate our approach for localisation on challenging radar data from the Oxford Radar RobotCar Dataset \cite{RadarRobotCarDatasetArXiv} through the tasks of odometry estimation and place recognition.

\subsection{Network Training}

We train our approach using 25 10km traversals from the Oxford Radar RobotCar Dataset \cite{RadarRobotCarDatasetArXiv} which provides Navtech CTS350-X radar data and ground truth radar poses. 
For training we convert the polar radar scan to Cartesian at either 0.7 or 0.35 m/pixel resolution and apply additional data augmentation to the odometry ground truth so that we can reliably solve pose at any rotation between radar scans.
For all training we use TensorFlow \cite{tensorflow2015-whitepaper} and the Adam \cite{kingma2014adam} optimiser with a learning rate of $\lambda = 10^{-3}$ until the task loss is minimised on a small validation set for at least 150k steps.

\subsection{Evaluation Metrics}

We evaluate our approach on 7 further dataset traversals for a total of approximately 70km. 

\subsubsection{Odometry}\label{sec:exp:odometry}
To quantify the performance of the odometry estimated by our point-based architecture, we compute translational and rotational drift rates using the approach proposed by the KITTI odometry benchmark \cite{geiger2013vision} for various resolutions and network configurations.
Specifically, we compute the average normalised end-point translational and rotational error for all subsequences of length (100, 200, \ldots , 800) metres compared to the ground truth radar odometry.

\subsubsection{Localisation}\label{sec:exp:localisation}
We follow the place recognition evaluation metrics as in \cite{arandjelovic2016netvlad,arandjelovic2014dislocation,gronat2013learning}. 
The query radar scan is deemed correctly localised if at least one of the top N retrieved radar scans, according to descriptor cosine distance, is within $d = 5m$ from the ground truth position of the query. 
For the purposes of this evaluation, detecting loop closures to the same trajectory are ignored and results are plotted for localising to other test datasets.
The metrics presented use $d = 5m$ rather than $d= 25m$ as in \cite{arandjelovic2016netvlad} because when nearer we can reliably solve for a full metric loop closure.

We further compare against the addition of a trainable NetVLAD layer \cite{arandjelovic2016netvlad} (512-D output / 64 clusters) to project location embeddings
(\cref{sec:learning:localisation}) 
onto a localisation specific metric space expecting this to improve performance. 
The highest performing model according to odometry metrics is frozen before the NetVLAD layer and fine-tuned for place recognition.
We use the batch hard triplet loss in \cref{alg:triplet_loss} with online hard negative mining, where $d(i,j)$ returns the $\ell_2$ distance between descriptors $i$ and $j$, sampling 5 positive locations ($\bm{p}$) closer than 5m and 5 negative locations ($\bm{n}$) further than 25m away per training sample. 

\vspace{-1.25em}
\begin{equation} \label{alg:triplet_loss}
\mathcal{L}_{place\ rec.} = max\left( \max_{p\in \bm{p}} d(a,p) - \min_{n\in \bm{n}} d(a,n) + m, 0 \right)
\end{equation}
\vspace{-.75em}

\newcommand{\SingleRowQuarter}{0.245\linewidth}
\newcommand{\TwoRowQuarter}{0.495\linewidth}

\begin{figure}{
    \includegraphics[width=\TwoRowQuarter]{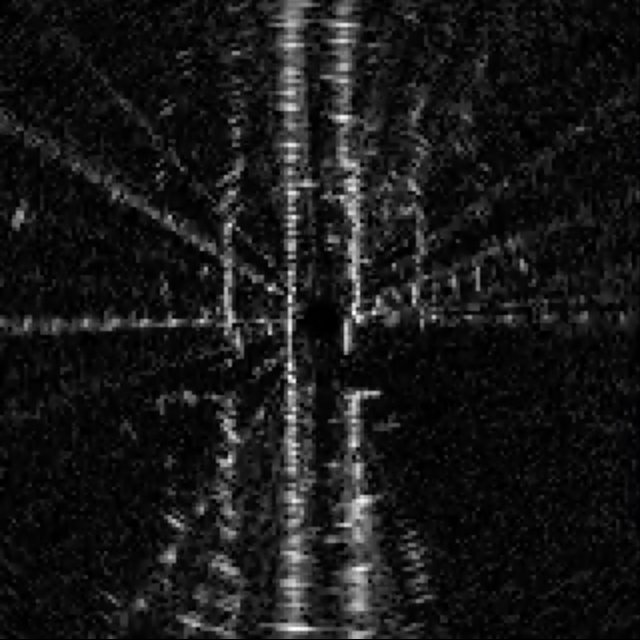}
    \hfill
    \includegraphics[width=\TwoRowQuarter]{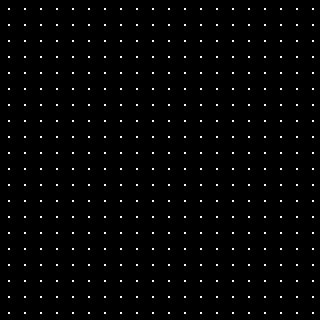}
    \\ \vspace{-1em} \\ 
    \includegraphics[width=\TwoRowQuarter]{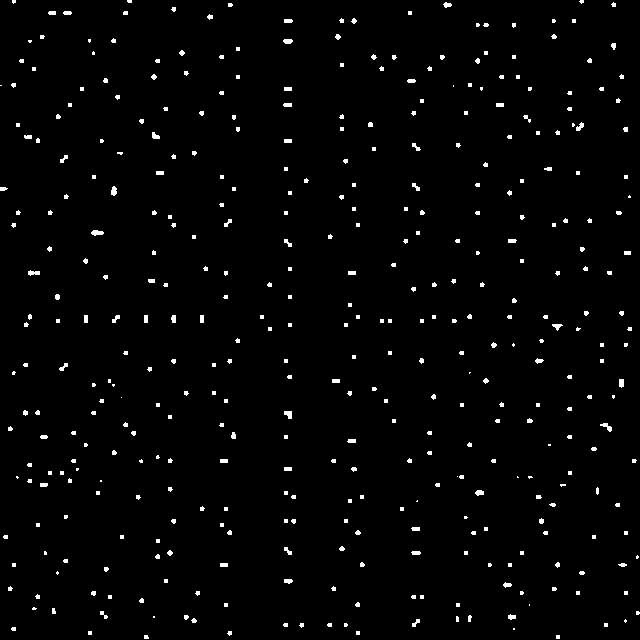}
    \hfill
    \includegraphics[width=\TwoRowQuarter]{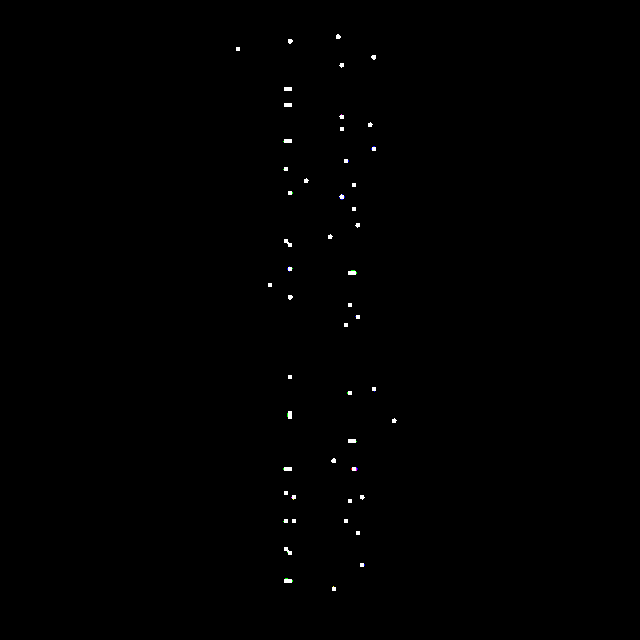}
    \caption{
    Architecture design experiments. 
    For a given radar input (top left) we compare against a baseline where points are distributed uniformly across the scan (top right) with Location and Score heads disabled. 
    With Location head enabled we learn to predict per cell sub-pixel keypoint locations (bottom left).
    When the Score head is also enabled (bottom right) we are able to ignore points due to noise artefacts or in unobservable regions, leaving only points located on the static structure in the scene. 
    }
    \label{fig:results:qualitative_network_predictions}}
\end{figure}

\begin{figure}

    \includegraphics[width=\TwoRowQuarter]{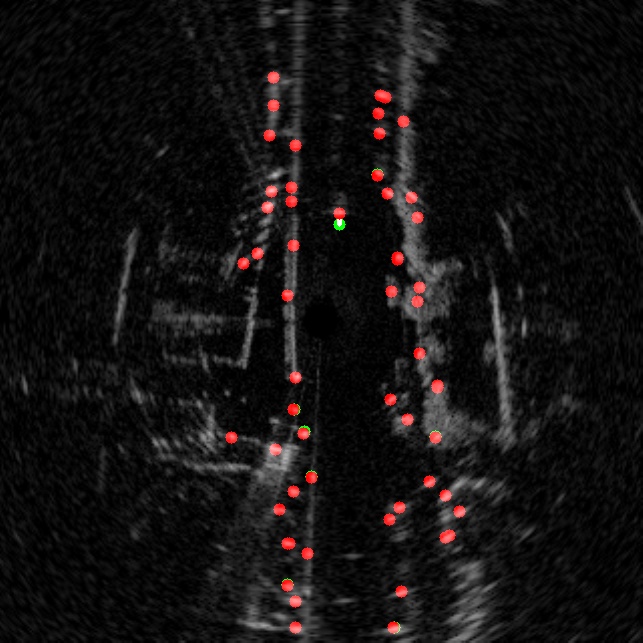}
    \hfill
    \includegraphics[width=\TwoRowQuarter]{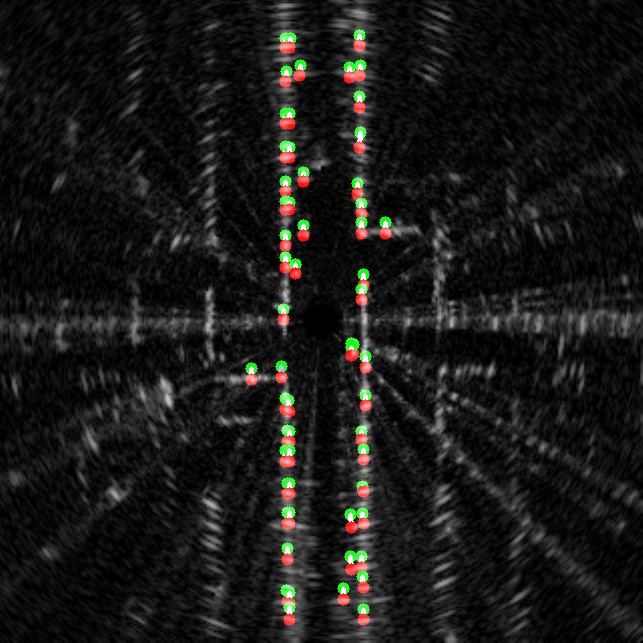}
    \\ \vspace{-1em} \\  
    \includegraphics[width=\TwoRowQuarter]{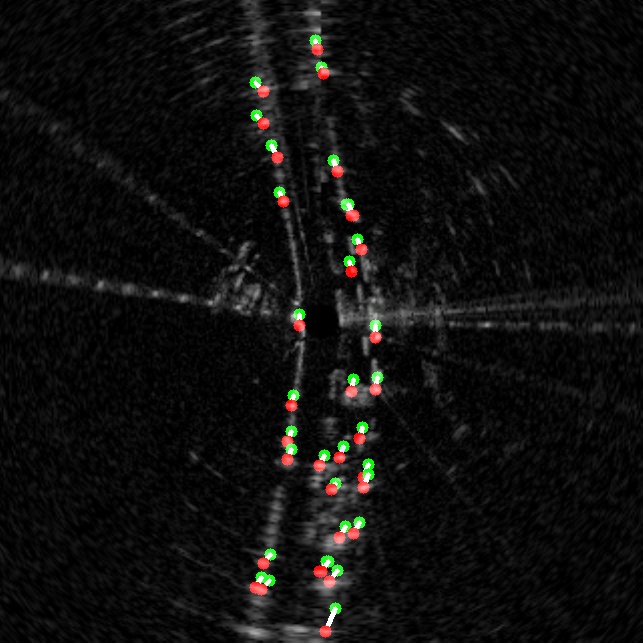}
    \hfill
    \includegraphics[width=\TwoRowQuarter]{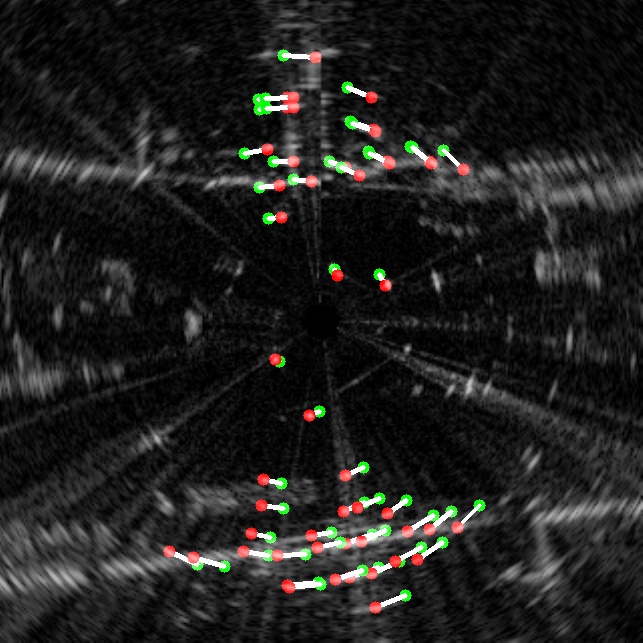}

    \caption{
    Odometry keypoint matches when stationary (top left), travelling forward (top right), on a slight bend (bottom left) and performing an aggressive turn (bottom right). 
    Points from sequential scans are shown in red or green and the match between them shown in white with only the highest scoring matches shown. 
    In all situations the point locations and weights accurately capture the vehicle motion and are well localised to the static structure in the scene such as walls and buildings. 
    Interestingly the only moving match in the stationary example belongs to a vehicle moving through the scene.
    }
    \label{fig:results:qualitative_matches}
\end{figure}


\sisetup{round-mode = places, round-precision = 4, group-separator='', detect-all}
\begin{table*}
    \centering
    \begin{tabular}{C{4.91cm}|c|c|c}
        \textbf{Benchmarks} & Translational Error (\%) & Rotational Error (deg/m) & Runtime (s) \\\hline
        RO Cen Full Res. \cite{cen2018precise} & \num{8.4730} & \num{0.0236} & \num{0.3059} \\
        RO Cen Equiv. \cite{cen2018precise} & \num{3.7168} & \num{0.0095} & \num{2.9036} \\
        CNN Regression \cite{li2018undeepvo} & \num{4.7683} & \num{0.0141} & \num{0.0060} \\
        Masking By Moving Equiv. * \cite{MaskingByMovingArXiv} & \num{1.5893} & \num{0.0044} & \num{0.0169} \\
        Stereo Visual Odometry * \cite{WinstonChurchill} & \num{3.9802} & \num{0.0102} & \num{0.0062} \\
    \end{tabular}
    \vspace{1.5em}

    \begin{tabular}{C{1.85cm}|C{1.05cm}|C{1.05cm}|c|c|c}
        \multicolumn{3}{c|}{\textbf{Ours}} & \multirow{2}{*}{Translational Error (\%)} & \multirow{2}{*}{Rotational Error (deg/m)} & \multirow{2}{*}{Runtime (s)}  \\  
        Resolution (m) & Localiser & Scores & & & \\\hline
        \multirow{4}{*}{\num{0.6912}} & & & \num{18.6996} & \num{0.0568853} & \num{0.01113330899} \\ 
        & \cmark & & \num{8.37003} & \num{0.0252867} & \num{0.01174369781} \\ 
        & & \cmark & \num{4.4153} & \num{0.0140383} & \num{0.01247057854} \\ 
        & \cmark & \cmark & \num{3.95176} & \num{0.0137888} & \num{0.01343412458} \\\hline 
        \multirow{4}{*}{\num{0.3456}} & & & \num{22.9889} & \num{0.0644324} & \num{0.0291232469} \\ 
        & \cmark & & \num{9.09554} & \num{0.0278042} & \num{0.03046170365} \\ 
        & & \cmark & \num{2.46071} & \num{0.00885271} & \num{0.03226511078} \\ 
        & \cmark & \cmark & \textbf{\num{2.05825}} & \textbf{\num{0.00672466}} & \num{0.0340129449300002} \\
    \end{tabular}
    \caption{
    Odometry drift evalutaion. Best performing model marked in bold.  Methods not directly comparable marked with *. 
    }
    \label{tab:results:odometry}
    \vspace{-2em}
\end{table*}

\section{Results}

\subsection{Odometry Performance}\label{sec:results:odometry}
The end-point-error evaluation is presented in \cref{tab:results:odometry}. 
The key benchmark we compare to is `RO Cen', the state-of-the-art in \emph{point-based} radar odometry, where `Full Res.' operates on the full resolution of the radar and `Equiv' is downsampled through max pooling to the resolution the algorithm was designed for. 
As can be seen, our best performing model (shown in bold) outperforms these by 45\% in translational and 29\% in rotational error whilst running an order of magnitude faster at 28.5Hz. 
Increasing the resolution would likely boost performance further at the cost of runtime speed.
Additionally we outperform pure CNN regression using a model designed and trained for pose estimation \cite{li2018undeepvo}. 

We provide two additional benchmarks not directly comparable to the method proposed. 
Firstly, odometry estimation in another modality using an off-the-shelf visual odometry system \cite{WinstonChurchill} 
as chosen by the prior state-of-the-art in radar odometry \cite{cen2018precise}, 
which we exceed in performance by a significant margin.
Secondly, the current state-of-the art in dense radar odometry estimation \cite{MaskingByMovingArXiv} at the most closely related configuration and resolution. 
Whilst we do not exceed the performance of \cite{MaskingByMovingArXiv}, we are not limited by rotation, crucial for solving metric loop closures in \cref{sec:results:mapping-and-localisation}.

We evaluate our architecture design optionally disabling the Location and Score heads. 
The effect these have on predicted keypoints are visualised in \cref{fig:results:qualitative_network_predictions}.
At both test resolutions, enabling both Location and Score heads lead to the best performance.
As the majority of a radar scan is either: empty, unobserved, or contain noise artefacts; scores prove more essential to odometry performance than locations as these regions can be ignored. 
Odometry keypoint matches are visualised \cref{fig:results:qualitative_matches} in various locations and vehicle movements, showing points localise well to the static structure.

\subsection{Localisation Performance}\label{sec:results:localisation-performance}

\begin{figure}
    \centering
    \includegraphics[width=\linewidth]{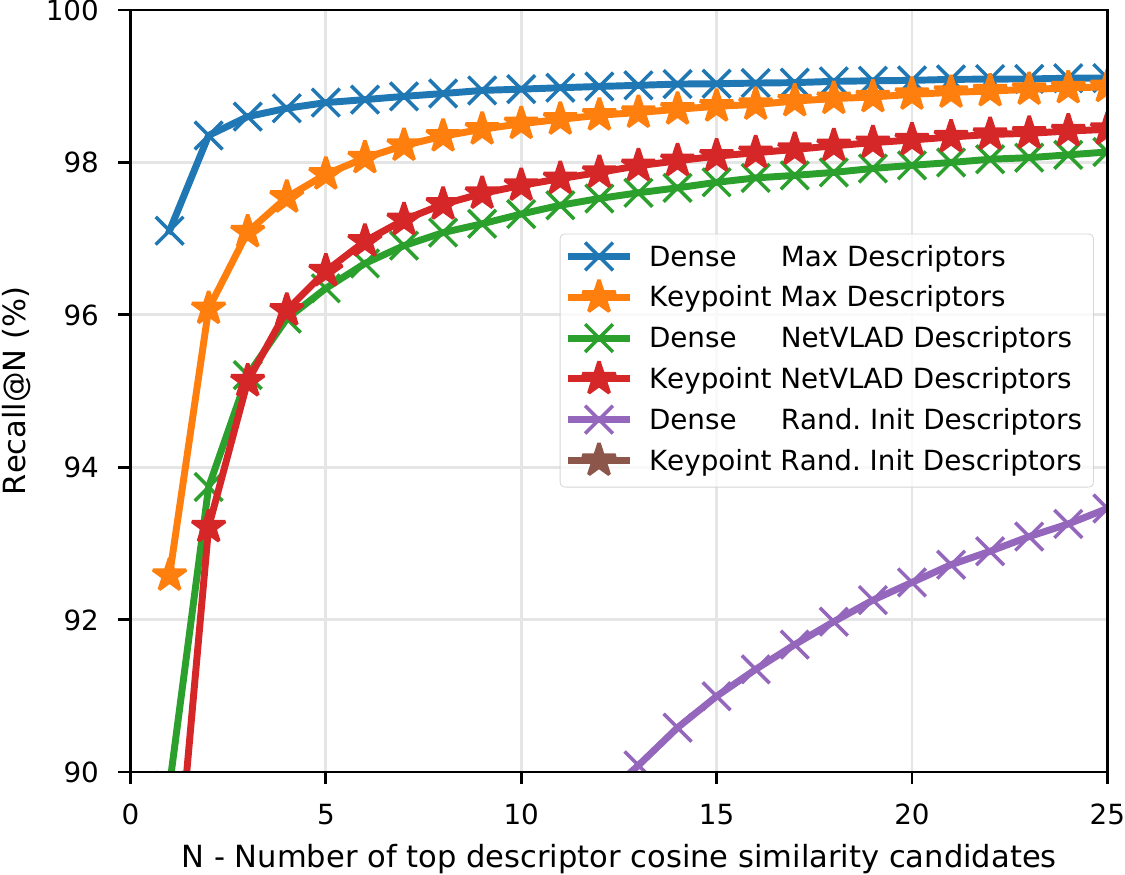}
    \caption{
    Place recognition results between test datasets showing Recall vs Number of top candidates as per the results in \cite{arandjelovic2016netvlad}. 
    The embeddings learnt by our architecture trained for odometry (`Max') exceed the performance of fine-tuning an additional layer to project onto a localisation specific metric space (`NetVLAD'). 
    }
    \label{fig:results:localisation_descriptor_performance}
    \vspace{-1.5em}
\end{figure}

Place recognition results are shown in \cref{fig:results:localisation_descriptor_performance}.
We compare creating location embeddings from the full resolution dense descriptor map and from descriptors extracted at keypoint locations.
Even when not trained on the task of place recognition, our location embeddings reliably allow us to detect topological loop closures (`Max Descriptors') far exceeding randomly initialised weights (`Rand.' with the keypoint variant off the bottom of the graph).

\begin{figure*}
    \centering
    \includegraphics[height=0.425\linewidth,clip,trim={0, 0, 32.5cm, 0}]{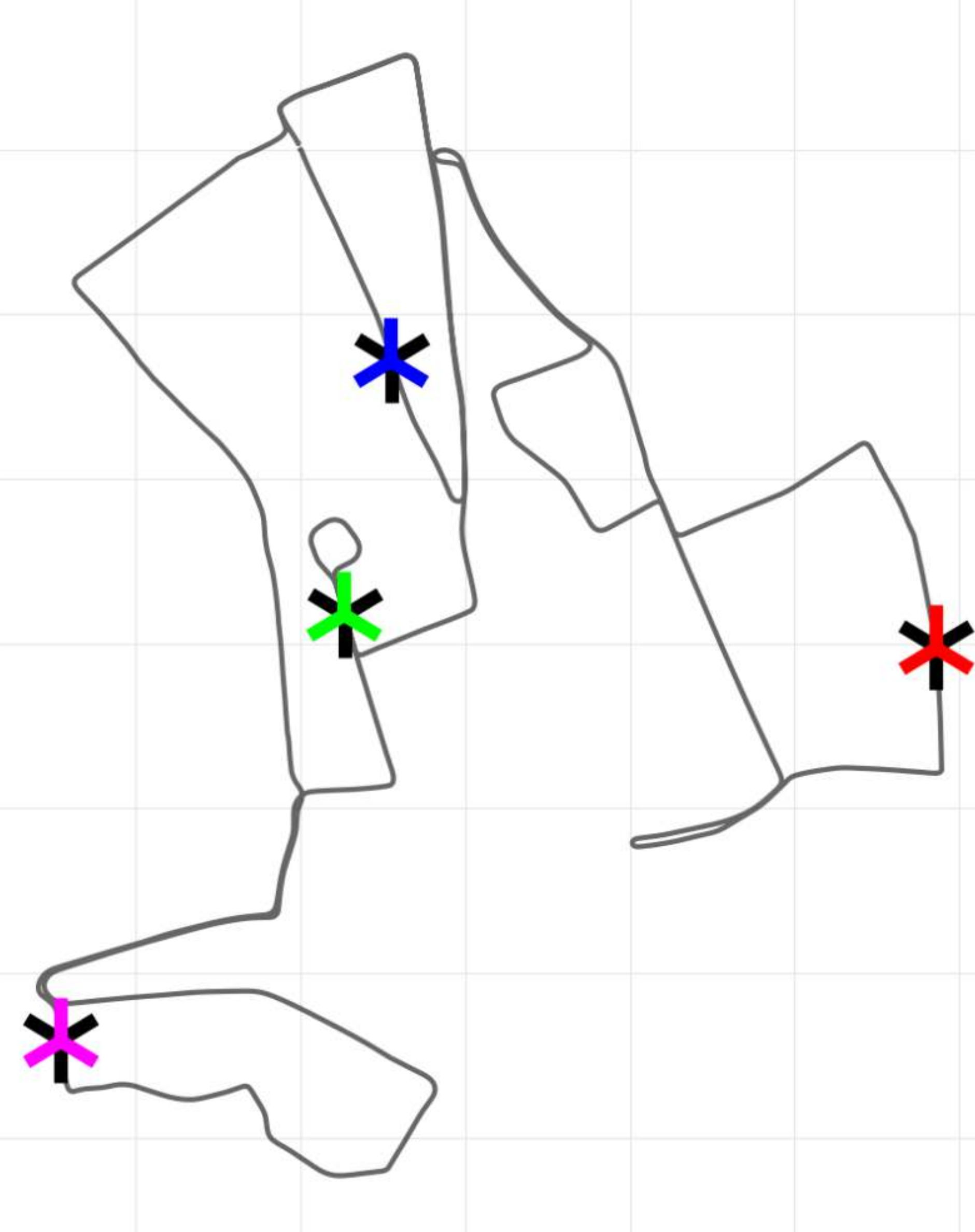}
    \hfill
    \includegraphics[height=0.425\linewidth,clip,trim={19.5cm, 0, 0, 0}]{figs/LoopClosureFigGs.pdf}
    \caption{
    Qualitative loop closure detections. For a given radar input, shown as 
    {\large$\pmb{\bm{\downY}}$}
    on the map (left) and top row (right), our location specific embeddings enable us to detect loop closures from different traversals of the route, shown as 
    \;\mbox{\large
    $\pmb{\mathrel{\raisebox{-0.1em}
    {\textcolor{red}{$\bm{\upY}$}\textcolor{blue}{$\bm{\upY}$}\textcolor{magenta}{$\bm{\upY}$}\textcolor{green}{$\bm{\upY}$}}
    }}$}
    on the map and the corresponding colour-coded scans in the third row. 
    As can be seen from the temporally closest camera images, place recognition can be extremely challenging in vision due to limited field-of-view, lens-glare and other environmental conditions.
    Using radar data, we are not faced with the same challenges. 
    }
    \label{fig:results:qualitative_loop_closures}
\end{figure*}

When fine tuning an additional layer for place recognition as described in \cref{sec:exp:localisation}, we freeze the best performing odometry estimation model (bottom row in \cref{tab:results:odometry}) before adding the NetVLAD layer.
Despite a better training convergence, interestingly the NetVLAD layer based embeddings generalise worse to the test set than the embeddings trained on the task of odometry. 
Further experiments increasing the dimensionality of the core architecture descriptors, as well as the NetVLAD layer, showed negligible improvements at the cost of runtime speed. Qualitative topological localisation results can be seen in \cref{fig:results:qualitative_loop_closures}.

\subsection{Mapping and Localisation}\label{sec:results:mapping-and-localisation}

Given we now have a system that can reliably solve the pose between two proximal radar scans and a method for detecting topological loop closures, we can combine these into a full mapping and localisation stack running at well over real-time speeds. 

\begin{figure*}
    \centering
    \includegraphics[height=0.335\linewidth,clip,trim={0, 0, 12.7cm, 0}]{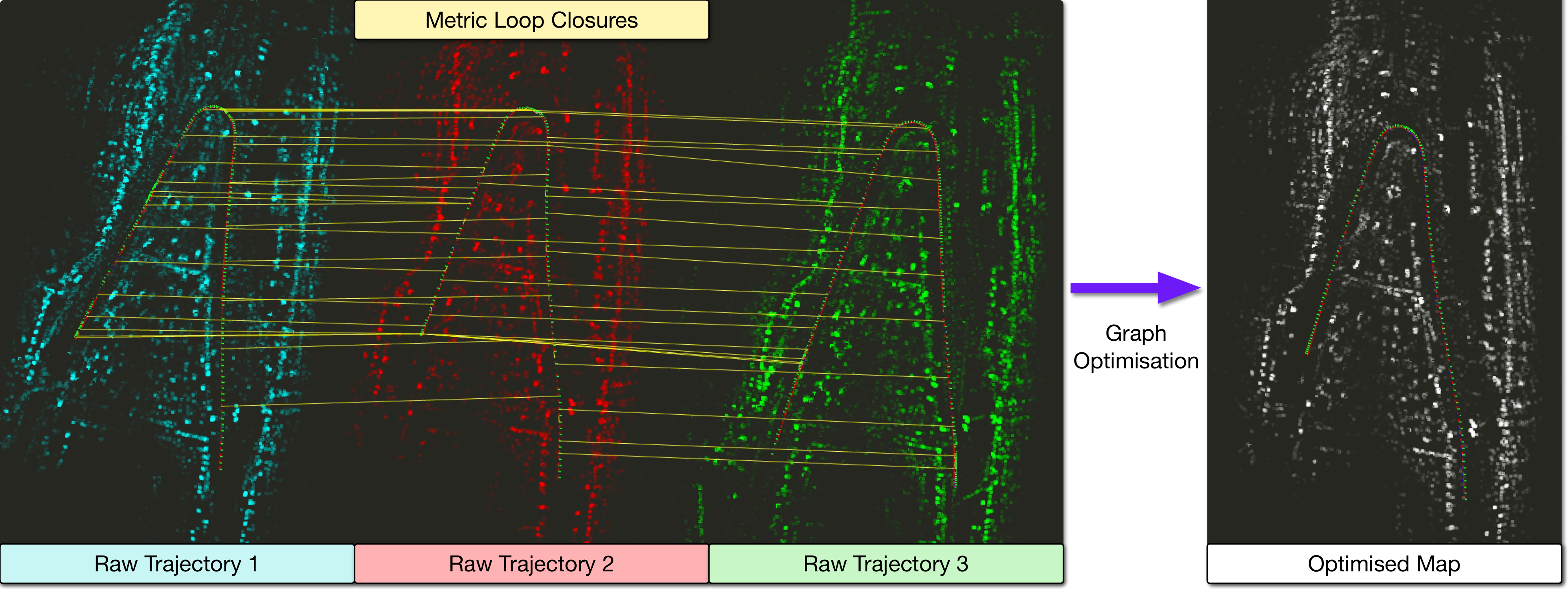}
    \hfill
    \includegraphics[height=0.335\linewidth,clip,trim={27.25cm, 0, 9.19cm, 0}]{figs/SLAM_figure_omnigraffle_thin.pdf}
    \hfill
    \includegraphics[height=0.335\linewidth,clip,trim={30.5cm, 0, 0, 0}]{figs/SLAM_figure_omnigraffle_thin.pdf}
    \caption{
    Full localisation and mapping system. 
    Given sequential radar frames we can estimate open loop trajectories by composing radar odometry and show three such sections from the test datasets on the left at approximately the same location. 
    All keypoints are rendered for each traversal weighted by keypoint scores (in cyan, red and green) and clearly highlight static structure, such as walls and buildings, whilst attenuating empty and unobserved regions. 
    When running our full system, as described in \cref{sec:learning:localisation}, we detect metric loop closures shown as yellow lines (downsampled heavily for visualisation). 
    All constraints are merged into a single map with pose graph optimisation, shown on the right in white, at well over real time speeds.
    }
    \label{fig:results:qualitative_slam}
\end{figure*}

For online applications we run three processes in parallel that output a fully optimised map. 
We run odometry estimation in a process to produce open loop trajectory edges. 
The second process detects topological loop closures by comparing against stored location embeddings, before solving the relative pose for metric loop closures. 
We store embeddings in an KDTree for fast lookup and set the cosine similarity threshold to give 100\% loop closure precision according to a small validation set.
The third process receives all edges and continuously optimises the underlying pose graph using g2o \cite{Kmmerle2011G2oAG}, producing a complete map of how the vehicle has travelled. 
A qualitative figure of our full mapping and localisation system can be seen in \cref{fig:results:qualitative_slam}.

\section{Conclusions}

In this paper we introduced the concept of learning keypoints for odometry estimation and localisation by embedding a differentiable pose estimator in our architecture. 
With this formulation, we learn to predict keypoint locations, scores and descriptors from pose information alone despite operating with extremely challenging radar data in complex environments.
Over a large test set we improve on the state-of-the-art in \emph{point-based} radar estimation by a large margin, reducing errors by up to 45\%, whilst running an order of magnitude faster. 
Whilst we do not surpass the current state-of-the-art in \emph{dense} radar odometry, we can solve poses at any rotation and detect metric loop closures, serving as a full system for radar-based mapping and localisation.

Furthermore, the benefits of our approach are not limited to radar or localisation tasks. 
The flexible architecture can be applied to most sensor modalities with few changes, and the detected points are readily reusable for other downstream tasks such as object velocity estimation. 
We plan to pursue these directions in the future, increasing radar based competencies for autonomous vehicles in urban environments.

\section{Acknowledgments}

This work was supported by the UK EPSRC Doctoral Training Partnership and EPSRC Programme Grant (EP/M019918/1). The authors would like to acknowledge the use of Hartree Centre resources in carrying out this work.


\cleardoublepage
\bibliographystyle{IEEEtran}
\bibliography{ms}

\end{document}